\documentclass[twoside]{article}
\usepackage[accepted]{aistats2012}

\usepackage{amssymb,amsthm}
\usepackage{color}
\usepackage{graphicx}
\usepackage{wrapfig}
\usepackage{amsmath,epsfig,subfigure}
\usepackage{cite}
\usepackage[authoryear,colon, sort&compress]{natbib}

\usepackage{enumerate}
\usepackage{enumitem}

\newcommand{\beq}{\vspace{0mm}\begin{equation}}
\newcommand{\eeq}{\vspace{0mm}\end{equation}}
\newcommand{\beqs}{\vspace{0mm}\begin{eqnarray}}
\newcommand{\eeqs}{\vspace{0mm}\end{eqnarray}}
\newcommand{\barr}{\begin{array}}
\newcommand{\earr}{\end{array}}

\newcommand{\Tmat}[0]{{{\bf T}}}

\newcommand{\Xmat}[0]{{{\bf X}}}
\newcommand{\Ymat}[0]{{{\bf Y}}}

\newcommand{\xv}{\boldsymbol{x}}
\newcommand{\yv}{\boldsymbol{y}}

\newcommand{\cdotv}{\boldsymbol{\cdot}}

\newcommand{\Thetamat}{\boldsymbol{\Theta}}

\newcommand{\Phimat}{\boldsymbol{\Phi}}

\newcommand{\thetav}{\boldsymbol{\theta}}

\newcommand{\phiv}{\boldsymbol{\phi}}

\newcommand{\R}{\mathbb{R}}
\newcommand{\E}{\mathbb{E}}
\newtheorem{thm}{Theorem}[section]

\newtheorem{lem}[thm]{Lemma}

\begin{document}


\runningauthor{Mingyuan Zhou, Lauren A. Hannah, David B. Dunson, Lawrence Carin}

\twocolumn[

\aistatstitle{Beta-Negative Binomial Process and Poisson Factor Analysis}

\aistatsauthor{Mingyuan Zhou \And Lauren A. Hannah$^\dag$ \And David B. Dunson$^\dag$ \And Lawrence Carin}

\aistatsaddress{Department of ECE, $^\dag$Department of Statistical Science, Duke University, Durham NC 27708, USA} ]

\begin{abstract}
A beta-negative binomial (BNB) process is proposed, leading to a beta-gamma-Poisson process, which may be viewed as a ``multi-scoop'' generalization of the beta-Bernoulli process. The BNB process is augmented into a beta-gamma-gamma-Poisson hierarchical structure, and applied as a nonparametric Bayesian prior for an infinite Poisson factor analysis model. A finite approximation for the beta process L\'{e}vy random measure is constructed for convenient implementation. Efficient MCMC computations are performed with data augmentation and marginalization techniques. Encouraging results are shown on document count matrix factorization.
\end{abstract}

\section{Introduction}

Count data appear in many settings. Problems include predicting future demand for medical care based on past use~\citep{CaTrMi88,DeTr97}, species sampling~\citep{ChristmasBirdData} and topic modeling of document corpora~\citep{LDA}.
Poisson and negative binomial distributions are typical choices for univariate and repeated measures count data; however, multivariate extensions incorporating latent variables (latent counts) are under developed. Latent variable models under the Gaussian assumption, such as principal component analysis and factor analysis, are widely used to discover low-dimensional data structure \citep{PPCA,GPLVM,West03bayesianfactor,Mingyuan09}. There has been some work on exponential family latent factor models that incorporate Gaussian latent variables \citep{DunsonDynamicLatentTrait,DunsonLVM03,GLTM,LVM_mixed}, but computation tends to be prohibitive in high-dimensional settings and the Gaussian assumption is restrictive for count data that are discrete and nonnegative, have limited ranges, and often present overdispersion.
In this paper we propose a flexible new nonparametric Bayesian prior to address these problems, the beta-negative binomial (BNB) process.

Using completely random measures \citep{Kingman}, \citet{JordanBP} generalize the beta process defined on $[0,1]\times\mathbb{R}^+$ by \cite{Hjort} to a general product space $[0,1]\times\Omega$, and define a Bernoulli process on an atomic beta process hazard measure to model binary outcomes. They further show that the beta-Bernoulli process is the underlying de Finetti mixing distribution for the Indian buffet process (IBP) of \citet{IBP}. To model count variables, we extend the measure space of the beta process to $[0,1]\times\mathbb{R}^+\times\Omega$ and introduce a negative binomial process, leading to the BNB process. We show that the BNB process can be augmented into a beta-gamma-Poisson process, and that this process may be interpreted in terms of a ``multi-scoop'' IBP. Specifically, each ``customer'' visits an infinite set of dishes on a buffet line, and rather than simply choosing to select certain dishes off the buffet (as in the IBP), the customer may select multiple scoops of each dish, with the number of scoops controlled by a negative binomial distribution with dish-dependent hyperparameters. As discussed below, the use of a negative binomial distribution for modeling the number of scoops is more general than using a Poisson distribution, as one may control both the mean and variance of the counts, and allow overdispersion. This representation is particularly useful for discrete latent variable models, where each latent feature is not simply present or absent, but contributes a distinct count to each observation.

We use the BNB process to construct an infinite discrete latent variable model called Poisson factor analysis (PFA), where an observed count is linked to its latent parameters with a Poisson distribution. To enhance model flexibility, we place a gamma prior on the Poisson rate parameter, leading to a negative binomial distribution.
The BNB process is formulated in a beta-gamma-gamma-Poisson hierarchical structure, with which we construct an infinite PFA model for count matrix factorization. We test PFA with various priors for document count matrix factorization, making connections to previous models; here a latent count assigned to a factor (topic) is the number of times that factor appears in the document.
%In particular we find that the BNB process PFA is well suited to topic modeling.

The contributions of this paper are: 1) an extension of the beta process to a marked space, to produce the beta-negative binomial (BNB) process; 2) efficient inference for the BNB process; and 3) a flexible model for count matrix factorization, which accurately captures topics with diverse characteristics when applied to topic modeling of document corpora.

\vspace{-1mm}
\section{Preliminaries}\label{sec:preliminaries}
\vspace{-1mm}
\subsection{Negative Binomial Distribution}

The Poisson distribution $X\sim \mathrm{Pois}(\lambda)$ is commonly used for modeling count data. It has the probability mass function $f_X(k) = { e^{\lambda}\lambda^k}/{k!}$, where $k \in\{0,1,\dots\}$, with both the mean and variance equal to $\lambda$. A gamma distribution with shape $r$ and scale $p/(1-p)$ can be placed as a prior on $\lambda$ to produce a negative binomial (a.k.a, gamma-Poisson) distribution as
\begin{align}\vspace{-1.05mm} \label{eq:gammaPoisson}
f_X(k) &= \int_0^\infty \mbox{Pois}(k;\lambda) \mbox{Gamma}\left(\lambda;r,p/(1-p)\right) d\lambda \nonumber\\
&=\frac{\Gamma(r+k)}{k!\Gamma(r)}(1-p)^r p^k \vspace{-1.05mm}
\end{align}
where $\Gamma(\cdot)$ denotes the gamma function. Parameterized by  $r>0$ and $p\in(0,1)$, this distribution  $X\sim \mbox{NB}(r,p)$  has a variance ${rp}/{(1-p)^2}$ larger than the mean ${rp}/(1-p)$, and thus it is usually favored %over the Poisson distribution
for modeling overdispersed count data. More detailed discussions about the negative binomial and related distributions and the corresponding stochastic processes defined on $\mathbb{R}^+$ can be found in \citet{NBLevy} and \citet{IntergerLevyProcess}.

\vspace{-1mm}
\subsection{L\'{e}vy Random Measures}
\vspace{-1mm}
Beta-negative binomial processes are created using L\'{e}vy random measures.
Following \citet{Wolp:Clyd:Tu:2011}, for any $\nu^+\ge0$ and any probability distribution $\pi(dp d\omega)$ on $\mathbb{R}\times\Omega$, let $K\sim \mbox{Pois}(\nu^+)$ and $\{(p_k,\omega_k)\}_{1\le k\le K} \stackrel{iid}{\sim} \pi(dp d\omega)$. Defining $\mathbf{1}_A(\omega_k)$ as being one if $\omega_k\in A$ and zero otherwise, the random measure  $\mathcal{L}(A)\equiv\sum_{k=1}^K \mathbf{1}_A(\omega_k)p_k$ assigns independent infinitely divisible random variables $\mathcal{L}(A_i)$ to disjoint Borel sets $A_i\subset\Omega$, with characteristic functions
\beq\label{LevyCF}
E\big[e^{it\mathcal{L}(A)}\big]=\exp\left\{\int\int_{\mathbb{R}\times A}(e^{itp}-1)\nu(dp d\omega)\right\}
\eeq
with $\nu(dp d\omega)\equiv \nu^+ \pi(dp d\omega)$. A random signed measure $\mathcal{L}$ satisfying (\ref{LevyCF}) is called a L\'{e}vy random measure. More generally, if the L\'{e}vy measure $\nu({dp d\omega})$  satisfies %the local $L_1$ integrability condition
\vspace{-1mm}
\beq \label{eq:LevyL1Condition}
\int\int_{\mathbb{R}\times S}(1\wedge |p|)\nu(dp d\omega) < \infty\vspace{-1mm}
\eeq
for each compact $S\subset \Omega$, it need not be finite for the L\'{e}vy random measure $\mathcal{L}$ to be well defined; the notation $1\wedge |p|$ denotes $\min\{1, |p|\}$. A nonnegative L\'{e}vy random measure $\mathcal{L}$ satisfying (\ref{eq:LevyL1Condition}) was called a completely random measure (CRM) by \citet{Kingman,PoissonP}  and an additive random measure by \citet{Ci11}. It was introduced for machine
learning by \citet{JordanBP} and \citet{JordanCRM}.

\vspace{-1mm}
 \subsection{Beta Process}\label{sec:preliminaries1}
 \vspace{-1mm}
 The beta process (BP) was defined by \citet{Hjort} for survival analysis with $\Omega = \R_+.$ \citet{JordanBP} generalized the process to an arbitrary measurable space $\Omega$ by defining a CRM $B$ on a product space $[0,1]\times\Omega$ with the L\'{e}vy measure
 \vspace{-1mm} \beq\label{eq:LevyBP}
 \nu_{\text{BP}}(dpd\omega) = cp^{-1}(1-p)^{c-1}dpB_0(d\omega).\vspace{-1mm}
 \eeq
 Here $c>0$ is a concentration parameter (or concentration function if $c$ is a function of $\omega$), $B_0$ is a  continuous finite measure over  $\Omega$, called the base measure, and $\alpha=B_0(\Omega)$ is the mass parameter. Since $\nu_{\text{BP}}(dp d\omega) $ integrates to infinity but satisfies (\ref{eq:LevyL1Condition}),
% \beq \int_{\Omega}\int_{0}^\infty \nu_{\text{BP}}(d\omega,dp)  = \infty,~~ \int_{\Omega}\int_{0}^\infty p\nu_{\text{BP}}(d\omega,dp)  < \infty \eeq
a countably infinite number of i.i.d. random points $\{(p_k,\omega_k)\}_{k=1,\infty}$ are obtained from the Poisson process with mean measure $\nu_{\text{BP}}$ and $\sum_{k=1}^{\infty}p_k$ is finite, where the atom $\omega_k \in \Omega$ and its weight $p_k\in [0,1]$.
Therefore, we can express a BP draw, $B\sim\mbox{BP}(c,B_0)$, as
$
 B=\sum_{k=1}^\infty p_k\delta_{\omega_k}
$, where $\delta_{\omega_k}$ is a unit measure at the atom $\omega_k$. If $B_0$ is discrete (atomic) and of the form
$B_0=\sum_k q_k\delta_{\omega_k}$, then
$B=\sum_k
p_k\delta_{\omega_k}$ with
$ p_k\sim\mbox{Beta}(cq_k,c(1-q_k))$.
If
$B_0$ is mixed discrete-continuous, $B$ is the sum of the two
independent contributions.

\vspace{-1mm}
\section{The Beta Process and the Negative Binomial Process}\label{sec:NB_beta}
\vspace{-1mm}
Let $B$ be a BP draw as defined in Sec. \ref{sec:preliminaries1}, and therefore $B=\sum_{k=1}^\infty p_k\delta_{\omega_k}$. A Bernoulli process $\mbox{BeP}(B)$ has atoms appearing at the same locations as those of $B$; it assigns atom $\omega_k$ unit mass with probability $p_k$, and zero mass with probability $1-p_k$, i.e., Bernoulli. Consequently, each draw from $\mbox{BeP}(B)$ selects a (finite) subset of the atoms in $B$. This construction is attractive because the beta distribution is a conjugate prior for the Bernoulli distribution. It has diverse applications including document classification \citep{JordanBP}, dictionary learning \citep{Mingyuan09, dHBP,BPFA2012} and topic modeling \citep{DictTopic}.

The beta distribution is also the conjugate prior for the negative binomial distribution parameter $p$, which suggests coupling the beta process with the negative binomial process. Further, for modeling flexibility it is also desirable to place a prior on the negative-binomial parameter $r$, this motivating a \emph{marked} beta process.

\vspace{-1mm}
\subsection{The Beta-Negative Binomial Process}\label{sec:BNB}
\vspace{-1mm}
Recall that a BP draw $B\sim\mbox{BP}(c,B_0)$ can be considered as a draw from the Poisson process with mean measure $\nu_{\text{BP}}$ in (\ref{eq:LevyBP}). We can mark a random point $(\omega_k,p_k)$ of $B$ with a random variable $r_k$ taking values in $\mathbb{R}^+$, where $r_k$ and $r_{k^{\prime}}$ are independent for $k\neq k^{\prime}$. Using the marked Poisson processes theorem~\citep{PoissonP}, $\{(p_k,r_k,\omega_k)\}_{k=1,\infty}$ can be regarded as random points drawn from a Poisson process in the product space $[0,1]\times\mathbb{R}^+\times\Omega$, with the L\'{e}vy measure
 \beq \label{eq:LevyMBP}
 \nu^*_{\text{BP}}(dpdr d\omega) =  cp^{-1}(1-p)^{c-1}dpR_0(dr)B_0(d\omega)
 \eeq
where $R_0$ is a continuous finite measure over $\mathbb{R}^+$ and $\gamma=R_0(\mathbb{R}^+)$ is the mass parameter. With (\ref{eq:LevyMBP}) and using $R_0 B_0$ as the base measure, we construct a marked beta process $B^*\sim \mbox{BP}(c,R_0 B_0)$, producing
\vspace{-1mm}
 \beq \label{eq:Bstar_base}
 B^*= \sum_{k=1}^{\infty}p_k\delta_{(r_k,\omega_k)}\vspace{-1mm}
 \eeq
where a point $(p_k,r_k,\omega_k)\sim \nu^*_{\text{BP}}(dpdr d\omega)/\nu^+_{\text{BP}}$ contains an atom $(r_k,\omega_k)\in \mathbb{R}^+\times \Omega$ with weight  $p_k\in[0,1]$.

With $B^*$ constituted as in (\ref{eq:Bstar_base}), we define the $i$th draw from a negative binomial process as $X_i\sim \mbox{NBP}(B^*)$, with
\beq
X_i = \sum_{k=1}^{\infty}\kappa_{ki} \delta_{\omega_k},~~~\kappa_{ki} \sim \mbox{NB}(r_k,p_k).
\eeq
The BP draw $B^*$ in (\ref{eq:Bstar_base}) defines a set of parameters $\{(p_k,r_k,\omega_k)\}_{k=1,\infty}$, and $(r_k,p_k)$ are used within the negative binomial distribution to draw a count $\kappa_{ki}$ for atom $\omega_k$. The $\{(p_k,r_k,\omega_k)\}_{k=1,\infty}$ are shared among all draws $\{X_i\}$, and therefore the atoms $\{\omega_k\}$ are shared; the count associated with a given atom $\omega_k$ is a function of index $i$, represented by $\kappa_{ki}$.

 In the beta-Bernoulli process \citep{JordanBP}, we also yield draws like $\{X_i\}$ above, except that in that case the $\kappa_{ki}$ is replaced by a one or zero, drawn from a Bernoulli distribution. The replacement of $\kappa_{ki}$ with a one or zero implies that in the beta-Bernoulli process a given atom $\omega_k$ is either used (weight one) or not (weight zero).
Since in the proposed beta-negative binomial process the $\kappa_{ki}$ corresponds to counts of ``dish'' $\omega_k$, with the number of counts drawn from a negative binomial distribution with parameters $(r_k,p_k)$, we may view the proposed model as a generalization to a ``multi-scoop'' version of the beta-Bernoulli process. Rather than simply randomly selecting dishes $\{\omega_k\}$ from a buffet, here each $X_i$ may draw \emph{multiple} ``scoops'' of each $\omega_k$.

\subsection{Model Properties}

Assume we already observe $\{X_i\}_{i=1,n}$. Since the beta and negative binomial processes are conjugate, the conditional posterior of $p_k$ at an observed point of discontinuity $(r_k,\omega_k)$ is
 \beq \label{eq:Bdiscrete}
p_k\sim\mbox{Beta}(m_{nk},c  + nr_k)
\eeq
where $m_{nk}= \sum_{i=1}^n\kappa_{ki}$ and $\kappa_{ki} = X_{i}\{\omega_k\}$. % and $c_{n}=c+m_{nk}+r_k$ are the total counts and concentration function, respectively.
The posterior of the L\'{e}vy measure of the continuous part
can be expressed as
\beq \label{eq:Bcontinuous}
\nu_{\text{BP},n}^*(dpdrd\omega) = cp^{-1}(1-p)^{c_n-1}dpR_0(dr)B_0(d\omega)
\eeq
where $c_n =c+nr$ is a concentration function. %We can understand this posterior as that an atom $\omega_{j^{\prime}}$, starting with the priors of a degenerate beta distribution on $p_k$ with parameter $c$ and ends with the posteriors of a degenerate beta distribution on $p_k$ with parameter $c+nr_{j^\prime}$ after seeing $n$ samples without observing a single count associated with $\omega_{j^{\prime}}$.
With (\ref{eq:Bdiscrete}) and (\ref{eq:Bcontinuous}), the posterior ${B^*}|\{X_{i}\}_{1,n}$ is defined, and following the notation in \citet{Kim, JordanBP,Thibaux,Miller}, it can be expressed as
\beq \label{eq:Bstar}
{B^*}|\{X_{i}\}_{1,n} \sim %\mbox{BP}\left(c_n,\frac{c}{c_n}R_0\times B_0 + \frac{1}{c_n}\sum_{i=1}^n X_i\right) =
\mbox{BP}\left(c_n, \frac{c}{c_n} R_0 B_0 + \frac{1}{c_n}\sum_{k}m_{nk}\delta_{(r_k,\omega_k)} \right)
\eeq
where
\begin{align}
%&B^*_n = \frac{c}{c_n} R_0 B_0 + \frac{1}{c_n}\sum_{k}m_{nk}\delta_{(r_k,\omega_k)}\\
c_n =
\begin{cases}
 c+m_{nk}+nr_k, & \text{if $(r,\omega) = (r_k,\omega_k) \in \mathcal{D}$}\\
 c+nr, & \text{if $(r,\omega) \in (\mathbb{R}^+\times \Omega)\backslash \mathcal{D}$}
\end{cases}
\end{align}
where $\mathcal{D}=\{(r_k,\omega_k)\}_k$ is the discrete space including all the points of discontinuity observed so far. % and  $(\mathbb{R}^+\times \Omega)\backslash \mathcal{D}$ is the space
\citet{JordanBP} showed that %under the assumption that the samples are exchangeable,
the IBP \citep{IBP} can be generated from the beta-Bernoulli process by marginalizing out the draw from the beta process. In the IBP metaphor, customer $n+1$ walks through a buffet line, trying a dish tasted by previous customers with probability proportional to the popularity of that dish among previous customers; additionally, this customer tries $K_{n+1}\sim\mbox{Pois}(\frac{c\alpha}{c+n})$ new dishes. By placing a count distribution on each observation, a BNB process prior naturally leads to a ``multiple-scoop'' generalization of the original IBP (i.e., a msIBP). That is, each customer now takes a number of scoops of each selected dish while walking through the buffet line.

Unlike the IBP, however, the msIBP predictive posterior distribution of $X_{n+1} | \{X_i\}_{i=1}^n$ does not have a closed form solution for a general measure $R_0$. This occurs because $R_0$ is not conjugate to the negative binomial distribution. We can analytically compute the distribution of the number of new dishes sampled. To find the number of new dishes sampled, note that an atom $\{(p_k,r_k,\omega_k)\}$ produces a dish that is sampled with probability $1-(1-p_k)^{r_k}$. Using the Poisson process decomposition theorem, we see that the number of sampled dishes has distribution $\mathrm{Pois}(\nu^+_{\text{msIBP}(n)})$,
\begin{align}\notag
 & \nu^+_{\text{msIBP}(n)} \\\notag
 %& = \alpha \int_{0}^{\infty} \int_{0}^1 c (1 - (1 - p)^r)p^{-1}(1-p)^{c_n-1}dp R_0(dr),\\\notag
 & =  \alpha \int_{0}^{\infty} \int_{0}^1 c (1 - (1 - p)^r)p^{-1}(1-p)^{c+nr-1}dp R_0(dr).
 \end{align}When $R_0 = \delta_{1}$, that is $r = 1$ and $\gamma = 1$, then the number of new dishes sampled has distribution $\mathrm{Pois}\left(\alpha \frac{c}{c+n}\right)$. Note that for choices of $R_0$ where $\nu^+_{\text{msIBP}(0)}$ is finite, $\nu^+_{\text{msIBP}(n)} \rightarrow 0$ as $n \rightarrow \infty$, and hence the number of new dishes goes to 0. Note that for the special case $R_0 = \delta_{1}$, the negative binomial process becomes the geometric process, and (\ref{eq:Bstar}) reduces to the posterior form discussed in \citet{Thibaux}; this simplification is not considered in the experiments, as it is often overly restrictive.

\vspace{-1mm}
\subsection{Finite Approximations for Beta Process}
\vspace{-1mm}
Since $\nu^+_\text{BP} = \nu^*_{\text{BP}}([0,1]\times\mathbb{R}^+\times\Omega)=\infty$, a BP generates countably infinite random points. For efficient computation, it is desirable to construct a finite L\'{e}vy measure which retains the random points of BP with non-negligible weights and whose infinite limit converges to $\nu^*_{\text{BP}}$. We propose such a finite L\'{e}vy measure as
\beq
\nu_{\epsilon\text{BP}}^*(dpdrd\omega) = cp^{c\epsilon-1}(1-p)^{c(1-\epsilon)-1}dpR_0(dr)B_0(d\omega)
\eeq
where $\epsilon>0$ is a small constant, and we have
\beq \label{eq:eBP+}
\nu^+_{\epsilon\text{BP}}=\nu^*_{\epsilon\text{BP}}([0,1]\times\mathbb{R}^+\times\Omega)=c\gamma\alpha\mathrm{B}(c\epsilon,c(1-\epsilon))
\eeq
where $\mathrm{B}(c\epsilon,c(1-\epsilon))$ is the beta function and it approaches infinity as $\epsilon\rightarrow 0$. Since
\beq
\lim_{\epsilon\rightarrow 0}\frac{\nu_{\epsilon\text{BP}}^*(dpdrd\omega)}{\nu_{\text{BP}}^*(dpdrd\omega)} = \lim_{\epsilon\rightarrow 0} {\left(\frac{p}{1-p}\right)}^{c\epsilon} = 1 \nonumber
\eeq
we can conclude that $\nu_{\epsilon\text{BP}}^*(dpdrd\omega)$ approaches $\nu_{\text{BP}}^*(dpdrd\omega)$ as $\epsilon\rightarrow 0$.

Using $\nu_{\epsilon\text{BP}}^*(dpdrd\omega)$ as the finite approximation, a draw from $\epsilon$BP $B^*_\epsilon\sim\epsilon\mbox{BP}(c,B_0)$ can be expressed as
\beq
B^*_\epsilon = \sum_{k=1}^K p_k\delta_{(r_k,\omega_k)},~~K\sim\mbox{Pois}(\nu^+_{\epsilon\text{BP}})\nonumber
\eeq
where $\{(p_k,r_k,\omega_k)\}_{k=1}^K \stackrel{iid}{\sim} \pi(dpdrd\omega)\equiv\nu^*_{\epsilon\text{BP}}(dpdrd\omega)/\nu^+_{\epsilon\text{BP}}$
and we have $\pi(dp)= \mbox{Beta}(p;c\epsilon,c(1-\epsilon))dp$, $\pi(dr)= R_0(dr)/\gamma$ and $\pi(d\omega)= B_0(d\omega)/\alpha$.
A reversible jump MCMC algorithm can be used to sample the varying dimensional parameter $K$. %Instead of letting $K$ be a random variable,
We can also simply choose a small enough $\epsilon$ and set $K = \mathbb{E}[\mbox{Pois}(\nu^+_{\epsilon\text{BP}})]= \nu^+_{\epsilon\text{BP}}.$

A finite approximation by restricting $p\in[\epsilon,1]$ \citep{Wolp:Clyd:Tu:2011} and stick breaking representations \citep{tehSB,StickBP} may also be employed to approximate the infinite model.

\section{Poisson Factor Analysis} \label{sec:PFA}

Given $K\le \infty$ and a count matrix $\Xmat \in \mathbb{R}^{P\times N}$ with $P$ terms and $N$ samples, discrete latent variable models assume that the entries of $\Xmat$ can be explained as a sum of smaller counts, each produced by a hidden factor, or in the case of topic modeling, a hidden topic. We can factorize $\Xmat$ under the Poisson likelihood as
\begin{equation}\label{eq:pfa}
\Xmat = \mbox{Pois}(\Phimat \Thetamat)
\end{equation}
where $\Phimat\in\mathbb{R}^{P\times K}$ is the factor loading matrix, each column of which is a factor encoding the relative importance of each term; $\Thetamat\in\mathbb{R}^{K\times N}$ is the factor score matrix, each column of which encodes the relative importance of each atom in a sample. %; and $K$ is the total number of atoms.
This is called Poisson factor analysis (PFA).

We can augment (\ref{eq:pfa}) as
\beq \label{eq:sumPois}
 x_{pi}   = \sum_{k=1}^K x_{pik},~x_{pik} \sim \mbox{Pois}( \phi_{pk} \theta_{ki})
\eeq
which is also used in \citet{Dunson05bayesianlatent} for a discrete latent variable model. This form is useful for inferring $\phi_{pk}$ and $\theta_{ki}$. As proved in Lemma \ref{lem:conditionalDist} (below), we can have another equivalent augmentation as
\begin{align}
&x_{pi} \sim \mbox{Pois} \left(\sum_{k=1}^K  \phi_{pk} \theta_{ki}\right),~\zeta_{pik} = \frac{\phi_{pk} \theta_{ki}}{\sum_{k=1}^K \phi_{pk} \theta_{ki}}\nonumber\\
&\label{eq:MultPois}[x_{pi1},\cdots,x_{piK}] \sim \mbox{Mult}\left(x_{pi}; \zeta_{pi1},\cdots,\zeta_{piK}\right)
\end{align}
which assigns $x_{pi}$ into the $K$ latent factors. Both augmentations in (\ref{eq:sumPois}) and (\ref{eq:MultPois}) are critical to derive efficient inferences, which were not fully exploited by related algorithms \citep{NMF,CannyGaP,DCA,FocusTopic}.

\begin{lem}\label{lem:conditionalDist}
Suppose that $x_1,\dots,x_K$ are independent random variables with
$x_k \sim \mathrm{Pois}(\lambda_k)$ and $x = \sum_{k=1}^K x_k$.
Set $\lambda = \sum_{k=1}^K\lambda_k$; let $(y,y_1,\dots,y_K)$ be random variables such that
\begin{align}\notag
y \sim \mathrm{Pois}(\lambda),~ (y_1,\dots,y_k) | y  \sim \mathrm{Mult}\left(y; \frac{\lambda_1}{\lambda},\dots,\frac{\lambda_K}{\lambda}\right).
\end{align}Then the distribution of $\xv=(x,x_1,\dots,x_K)$ is the same as the distribution of $\yv=(y,y_1,\dots,y_K)$.
\end{lem}

\vspace{-3mm}
\begin{proof} For $t=[t_0,\cdots,t_K] \in \R^{K+1}$ and compare the characteristic functions (CF)  of $\xv$ and $\yv$,
%\begin{align}\notag
\begin{align}
\E\left[ e^{it^T\xv}\right]  &\hspace{-0.8mm}=\hspace{-0.8mm} \prod_{k=1}^K \E \left[e^{it(t_0+t_{k})x_k}\right] \hspace{-0.8mm}= \hspace{-0.8mm}e^{\sum_{k=1}^K \lambda_k e^{i(t_0+t_{k})-1} };\nonumber\\
\E \left[ e^{it^T\yv} \right] & \hspace{-0.8mm} = \hspace{-0.8mm} \E \left[ \E \left[ \left. e^{it^T\yv}\right| y \right]\right] \hspace{-0.8mm}=\hspace{-0.8mm} \E \left[ \left( \sum_{k=1}^K \frac{\lambda_k}{\lambda} e^{i(t_0+t_k)}\right)^y\right]\notag\\
 &\hspace{-0.8mm}= \hspace{-0.4mm}e^{\sum_{k=1}^K \lambda_k e^{i(t_0+t_{k})-1} }.\notag
\end{align}
Since the CF uniquely characterizes the distribution, the distributions are the same.\end{proof}

\vspace{-3mm}
\subsection{Beta-Gamma-Gamma-Poisson Model}
\vspace{-3mm}

Notice that for a sample, the total counts are usually observed, and it is the counts assigned to each factor that are often latent and need to be inferred. However, the BNB process in Section \ref{sec:BNB} does not tell us how the total counts are assigned to the latent factors.
Recalling (\ref{eq:gammaPoisson}), the negative binomial distribution is a gamma-Poisson mixture distribution. Therefore, we can equivalently represent $X_i\sim \mbox{NBP}(B^*)$ as
$ X_i\sim \mathcal{P}({\mathcal{T}({B^*})}) $, where $\mathcal{T}({B^*})$ is a gamma process defined on $B^*$ and $\mathcal{P}({\mathcal{T}({B^*})})$ is a Poisson process defined on $\mathcal{T}({B^*})$.
Thus at atom $\omega_k$, the associated count $\kappa_{ki} \sim \mbox{NB}(r_k,p_k)$ can be expressed as
\beq
\kappa_{ki} \sim \mbox{Pois}(\theta_{ki}),~~\theta_{ki}\sim \mbox{Gamma}(r_k,{p_k}/{(1-p_k)})% \\
%\kappa_k \sim  \int_{0}^\infty \mbox{Pois}(\kappa_k; \theta_k)\mbox{Gamma}\left(\theta_k; r_k,{p_k}/{(1-p_k})\right) d\theta_k\nonumber
\eeq
where $\theta_{ki}$ is the weight of $X_i$ at atom $\omega_k$ in the gamma process $\mathcal{T}({B^*})$. If we further place a gamma prior on $r_k$, then $(p_k,r_k,\theta_{ki},\kappa_{ki})$ naturally forms a beta-gamma-gamma-Poisson hierarchical structure, and we call the BNB process PFA formulated under this structure the $\beta\gamma\Gamma$-PFA.  This kind of hierarchical structure is useful for sharing statistical strength between different groups of data, and efficient inference is obtained by exploiting conjugacies in both the beta-negative binomial and gamma-Poisson constructions.

Using $\epsilon$BP $\nu_{\epsilon\text{BP}}^*(dpdrd\phiv)$ on $[0,1]\times\mathbb{R}^+\times \mathbb{R}^{P}$ as the base measure for the negative binomial process, we construct the $\epsilon$BNB process and apply it as the nonparametric Bayesian prior for PFA. We have $K\sim\mbox{Pois}({\nu_{\epsilon\text{BP}}^{+}})=\mbox{Pois}(c\gamma\alpha\mathrm{B}(c\epsilon,c(1-\epsilon)))$, and thus $K\le \infty$ with the equivalence obtained at $\epsilon=0$. Using the beta-gamma-gamma-Poisson construction, we have
\begin{align}
%x_{pi} &   \sim \mbox{Pois}\left( \sum_{k=1}^K \phi_{pk} \theta_{ki}\right),~K\sim\mbox{Pois}({\nu_{\epsilon\text{BP}}^{+}})\\
\label{eq:Augment1}
x_{pi} &  = \sum_{k=1}^K x_{pik},~x_{pik} \sim \mbox{Pois}( \phi_{pk} \theta_{ki})\\
\phiv_{k} &\sim \mbox{Dir}\left(a_{\phi},\cdots,a_{\phi}\right)\\
\label{eq:theta}
\theta_{ki} &\sim \mbox{Gamma}\left(r_k,\frac{p_k}{1-p_k}\right)\\
r_k &\sim \mbox{Gamma}(c_0 r_0, 1/c_0)\\
\label{eq:p_k}
p_k  &\sim \mbox{Beta}(c\epsilon,c(1-\epsilon)).
\end{align}

\subsection{MCMC Inference}
\vspace{-1mm}
Denote $x_{ \cdotv ik} = \sum_{p=1}^P x_{pik}$, $x_{p \cdotv k} = \sum_{i=1}^N x_{pik}$, $x_{\cdotv \cdotv k} = \sum_{p=1}^P \sum_{i=1}^N x_{pik}$ and  $x_{\cdotv i \cdotv} = \sum_{p=1}^P \sum_{k=1}^K x_{pik}$. The size of $K$ is upper bounded by ${\nu_{\epsilon\text{BP}}^{+}}=c\gamma\alpha\mathrm{B}(c\epsilon,c(1-\epsilon))$.

\textbf{\emph{Sampling}} $x_{pik}$. Use (\ref{eq:MultPois}).

\textbf{\emph{Sampling} $\phiv_{k}$}.
Exploiting the relationships between the Poisson and multinomial distributions and using $\sum_{p=1}^{P}\phi_{pk} = 1$, one can show that
$%\begin{align}\label{eq:MultPois1}
p([x_{1ik},\cdots,x_{Pik}]|-) \sim \mbox{Mult}\left(x_{\cdotv ik};\phiv_k\right) \nonumber
$, %\end{align}
thus we have
\begin{align}\label{eq:samplephiv}
%P(\phi_{pk}|-) &\sim \mbox{Gamma}(a_{\phi}/P + x_{p\cdotv k},b_{\phi} + \sum_{i=1}^N\theta_{ki})\\
p(\phiv_k|-)&\sim \mbox{Dir}\left(a_{\phi} + x_{1\cdotv k},\cdots,a_{\phi} + x_{P \cdotv k}\right).
\end{align}
\textbf{\emph{Sampling} $p_{k}$}. Marginalizing $\phiv_{k}$  and $\theta_{ki}$ out,
$%\begin{align}
x_{\cdotv ik} \sim \mbox{NB}( r_k, p_k),~~ p_k  \sim \mbox{Beta}(c\epsilon,c(1-\epsilon)) \nonumber
$, %\end{align}
thus
\begin{align}\label{eq:p_k_post}
p(p_{k}|-)  &\sim \mbox{Beta}(c\epsilon + x_{\cdotv \cdotv k},c(1-\epsilon)+ Nr_k).
\end{align}
\textbf{\emph{Sampling}} $r_k$. It can be shown that
$%\beq
p(r_k|-) \propto \mbox{Gamma} (r_k; c_0r_0,1/c_0) \prod_{i=1}^N \mbox{NB}\left(x_{\cdotv i k}; r_k,p_k\right)\nonumber
$, %\eeq
thus
\beq\label{eq:r_k0} p(r_k|-) \sim \mbox{Gamma}\left(c_0r_0, \frac{1}{c_0-N\log(1-p_k)}\right) \eeq
if $x_{\cdotv \cdotv k}=0$. If $x_{\cdotv \cdotv k}\neq0$, we prove in Lemma \ref{lem:gConcave} that $g(r_k) = \log p(r_k|-)$ is strictly concave if $c_0r_0\ge1$,
then we can use Newton's method to find an estimate as
\beq \tilde{r}_k = r_k^t - \frac{p^{\prime}(r_k^t|-) }{p^{\prime\prime}(r_k^t|-)}
\nonumber \eeq
which can be used to construct a proposal in a Metropolis-Hastings (MH) algorithm as
\beq\label{eq:randomMH} r_k^\prime \sim \mathcal{N}\left(\tilde{r}_k, \mu \sqrt{\tilde{r}_k}\right) \eeq
where $\mu$ is an adjustable stepsize.
Note that the adaptive rejection sampling in \cite{AdapRejection} may also be used to sample $r_k$.

\textbf{\emph{Sampling} $\theta_{ki}$}. Using (\ref{eq:Augment1}) and (\ref{eq:theta}), we have%Due to the conjugacy between Poisson and gamma distributions, we have
\begin{align}\label{eq:SampleTheta}
p(\theta_{ki}|-) &\sim \mbox{Gamma}(r_k + x_{\cdotv ik}, p_k).
%P(b_s|-) &\sim \mbox{Gamma}(e_0 + NKa_{s},f_0 + \sum_{i=1}^N\sum_{k=1}^K s_{ki})
\end{align}

\begin{lem}\label{lem:gConcave}
If $x_{\cdotv \cdotv k}\neq0$, then for any $c_0r_0\ge1$, $g(r_k) = \log p(r_k|-)$ is strictly concave.
\end{lem}

\vspace{-3mm}
\begin{proof}
Since $ %\begin{align}
g^{\prime}(r_k)  = (c_0r_0-1)/r_k -c_0 + N\log(1-p_k)- N\psi(r_k) + \sum_{i=1}^N \psi(r_k+x_{\cdotv i k}) \nonumber
$ and $
g^{\prime\prime}(r_k)  = - (c_0r_0-1)/r_k^2 - N \psi_1(r_k) + \sum_{i=1}^N \psi_1(r_k+x_{\cdotv i k})\nonumber
$, %\end{align}
where $\psi(x)$ is the diagmma function and $\psi_1(x)$ is the trigamma function which is strictly decreasing for $x>0$, if $c_0 r_0\ge1$ and $x_{\cdotv \cdotv k}\neq0$, we have $g^{\prime\prime}(r_k)<0$, and thus $g(r_k)$ is strictly concave and has a unique maximum.
\end{proof}

\vspace{-3mm}
\section{Related Discrete Latent Variable Models}\label{Sec:RelatedModel}
\vspace{-2mm}
The hierarchical form of the $\beta\gamma\Gamma$-PFA  model shown in (\ref{eq:Augment1})-(\ref{eq:p_k}) can be modified in various ways to connect to previous discrete latent variable models. For example, we can let $\{\theta_{ki}\}_{i=1,N}\equiv g_k$ and $g_k\sim\mbox{Gamma}(g_0/K,1)$, resulting in the infinite gamma-Poisson feature model in \citet{InfGaP} as $K\rightarrow \infty$.  \citet{Thibaux} showed that it can also be derived from a gamma-Poisson process. Although this is a nonparametric model supporting an infinite number of  features, requiring $\{\theta_{ki}\}_{i=1,N}\equiv g_k$ may be too restrictive. We mention that \citet{NBPJordan} have independently investigated beta-negative binomial processes for mixture and admixture models.

Before examining the details, defined by $p_k$ and  $r_k$ in (\ref{eq:theta}), $z_{ki}$ in (\ref{eq:z_ki}) and the subset of the $K$ factors needed to represent the data are inferred, we summarize in Table \ref{Tab:Relationships} the connections between related algorithms and PFA with various priors, including non-negative matrix factorization (NMF) \citep{NMF}, latent Dirichlet allocation (LDA) \citep{LDA}, gamma-Poisson (GaP) \citep{CannyGaP} and the focused topic model (FTM) \citep{FocusTopic}.
Note that the gamma scale parameters $p_k/(1-p_k)$ and $p_{k^\prime}/(1-p_{k^\prime})$ in (\ref{eq:theta}) are generally different for $k\neq k^{\prime}$ in $\beta\gamma\Gamma$-PFA, and thus the normalized factor score ${\tilde{\thetav}_i}={\thetav_i}\big/{\sum_{k=1}^K \theta_{ki}}$ does not follow a Dirichlet distribution. This is a characteristic that distinguishes $\beta\gamma\Gamma$-PFA from models with a global gamma scale parameter, where the normalized factor scores follow Dirichlet distributions.

\begin{table}\caption{\small Algorithms related to PFA under various prior settings. The gamma scale and shape parameters and the sparsity of the factor scores are controlled by $p_k$, $r_k$ and $z_{ki}$, respectively.}
\centering
{\small
\begin{tabular}{|c|c|c|c|c|c|}
  \hline
  PFA                       & Infer & Infer & Infer &Infer & Related  \\
  priors                       & $p_k$ & $r_k$ & $z_{ki}$ &$K$& algorithms \\ \hline
  $\Gamma$             & $\times$ & $\times$ & $\times$ & $\times$ & NMF \\
  Dir                   & $\times$ & $\times$ & $\times$ & $\times$ & LDA \\
  $\beta\Gamma$        & $\checkmark$ & $\times$ & $\times$& $\checkmark$& GaP \\
%  $\gamma\Gamma$        & $\times$ & $\checkmark$ & $\times$ & HGaP\\
  S$\gamma\Gamma$       & $\times$ & $\checkmark$ & $\checkmark$& $\checkmark$& FTM\\
  $\beta\gamma\Gamma$   & $\checkmark$ & $\checkmark$ & $\times$& $\checkmark$& \\
  \hline
\end{tabular}\label{Tab:Relationships}
}\vspace{-2mm}
\end{table}

\vspace{-1.5mm}
\subsection{Nonnegative Matrix Factorization and a Gamma-Poisson Factor Model}
\vspace{-1mm}
We can modify the $\beta\gamma\Gamma$-PFA into a $\Gamma$-PFA by letting \begin{align}
\label{eq:gPFA1}\phi_{pk} &\sim \mbox{Gamma}(a_{\phi},1/b_{\phi})\\
\label{eq:gPFA2}\theta_{ki} &\sim \mbox{Gamma}(a_{\theta},g_k/a_{\theta}).\end{align} Using (\ref{eq:Augment1}), (\ref{eq:gPFA1}) and (\ref{eq:gPFA2}), one can show that
\begin{align}\vspace{-2mm}\label{eq:gamma1}
\hspace{-3mm} p(\phi_{pk}|-) &\sim \mbox{Gamma}(a_{\phi} +  x_{p\cdotv k},1/(b_\phi + \theta_{k\cdotv}) )
\\
\label{eq:gamma2}
 \hspace{-3mm} p(\theta_{ki}|-) &\sim \mbox{Gamma}(a_{\theta} + x_{\cdotv ik},1/(a_{\theta}/g_k + \phi_{\cdotv k}))\vspace{-2mm}
\end{align}
where $\theta_{k\cdotv} = \sum_{i=1}^N\theta_{ki}$  and $\phi_{\cdotv k}  = \sum_{p=1}^P\phi_{pk}$. % and $p((x_{pi1},\cdots,x_{piK})|-)$ is shown in (\ref{eq:MultPois}).
If $a_{\phi}\ge 1$ and $a_{\theta}\ge 1$, using (\ref{eq:MultPois}), (\ref{eq:gamma1}) and (\ref{eq:gamma2}), we can substitute $\mathbb{E}[x_{pik}]$ into the modes of $\phi_{pk}$ and $\theta_{ki}$, leading to an Expectation-Maximization (EM) algorithm as
\begin{align}\label{eq:NMF1}
 \phi_{pk} &= \phi_{pk} \frac{\frac{a_\phi-1}{\phi_{pk} }+\sum_{i=1}^N \frac{x_{pi}\theta_{ki}}{\sum_{k=1}^K \phi_{pk}\theta_{ki}}}{b_\phi+\theta_{k\cdotv} } \\
 \label{eq:NMF2}
 \theta_{ki} &= \theta_{ki} \frac{ \frac{a_\theta-1}{\theta_{ki} } + \sum_{p=1}^P \frac{x_{pi}\phi_{pk}}{\sum_{k=1}^K\phi_{pk}\theta_{ki}}}{a_{\theta}/g_k + \phi_{\cdotv k}}.
 \end{align}
If we set $b_{\phi}=0$, $a_{\phi}=a_{\theta}=1$ and $g_k=\infty$, then  (\ref{eq:NMF1}) and (\ref{eq:NMF2})
 are the same as those of non-negative matrix factorization (NMF) with an objective function of minimizing the KL divergence $D_{KL}(\Xmat||\Phimat\Thetamat)$~\citep{NMF}. If we set $b_{\phi}=0$ and $a_{\phi}=1$, then  (\ref{eq:NMF1}) and (\ref{eq:NMF2})
 are the same as those of the gamma-Poisson (GaP) model of \citet{CannyGaP}, in which setting $a_\theta=1.1$ and estimating $g_k$ with $g_k=\mathbb{E}[\theta_{ki}]$ are suggested. Therefore, as summarized in Table \ref{Tab:Relationships}, NMF is a special cases of the $\Gamma$-PFA, which itself can be considered as a special case of the $\beta\gamma\Gamma$-PFA with fixed $r_k$ and $p_k$. If we impose a Dirichlet prior on $\phiv_{k}$, the GaP can be consider as a special case of $\beta\Gamma$-PFA, which itself is a special case of the $\beta\gamma\Gamma$-PFA with a fixed $r_k$.

\subsection{Latent Dirichlet Allocation}
We can modify the $\beta\gamma\Gamma$-PFA into a Dirichlet PFA (Dir-PFA) by changing the prior of $\thetav_i$ to
$
\thetav_{i} \sim \mbox{Dir}\left(a_{\theta},\cdots,a_{\theta}\right)$. Similar to the derivation of $p(\phiv_k|-)$ in (\ref{eq:samplephiv}), one can show $p([x_{\cdotv i1},\cdots,x_{\cdotv iK}]|-)
\sim \mbox{Mult}\left(x_{\cdotv i \cdotv}; \thetav_i\right)$ and thus
$p(\thetav_i|-)\sim \mbox{Dir}\left( a_{\theta} + x_{\cdotv i1},\cdots,a_{\theta} + x_{\cdotv iK}\right)$.
Dir-PFA and LDA \citep{LDA,OnlineLDA} have the same block Gibbs sampling and variational Bayes inference equations (not shown here for brevity). It may appear that Dir-PFA should differ from LDA via the Poisson distribution; however, imposing Dirichlet priors on both factor loadings and scores makes it essentially lose that distinction.

\subsection{Focused Topic Model}
We can construct a sparse $\gamma\Gamma$-PFA (S$\gamma\Gamma$-PFA) with the beta-Bernoulli process prior by letting $\theta_{ki} = z_{ki}s_{ki}$ and
\begin{align} %\label{eq:PisonLDA}
%\phiv_{k} &\sim \mbox{Dir}\left(a_{\phi}/P,\cdots,a_{\phi}/P\right)\\
s_{ki}&\sim \mbox{Gamma}(r_k,p_k/(1-p_k)),~r_k \sim \mbox{Gamma}(r_0,1)\nonumber\\
\label{eq:z_ki}z_{ki} &\sim \mbox{Bernoulli}(\pi_k),~\pi_k \sim \mbox{Beta}(c\epsilon,c(1-\epsilon)).
\end{align}
If we fix $p_k=0.5$, it can be shown that under the PFA framework,  conditioning on $z_{ki}$, we have
\begin{align} \label{eq:FTM1} x_{\cdotv i k}  &\sim
%\delta(z_{ki}-1)\int_0^\infty \mbox{Pois}(x_{\cdotv i k};z_{ki} %s_{ki})\mbox{Gamma}(s_{ki};\alpha_k,\frac{0.5}{1-0.5}) d s_{ki}=
\mbox{NB}\left(z_{ki}r_k,0.5\right) \\
 \label{eq:FTM2}
x_{\cdotv i \cdotv}  &\sim \mbox{NB}\left(\sum_{k=1}^K z_{ki}r_k,0.5\right) \\
\label{eq:FTM3}    \tilde{\thetav}_{i}&= {\thetav_i}\bigg/{\sum_{k=1}^K \theta_{ki}} \sim \mbox{Dir}(z_{1i}r_1, \cdots,z_{Ki}r_K)\\
%&x_{pi} \sim \mbox{Pois} \left(\sum_{k=1}^K  \phi_{pk} \theta_{ki}\right),~\zeta_{pik} = \frac{\phi_{pk} \theta_{ki}}{\sum_{k=1}^K \phi_{pk} \theta_{ki}}\nonumber\\
[x_{pi1}&,\cdots,x_{piK}] \sim \mbox{Mult}\left(x_{pi}; \tilde{\zeta}_{pi1},\cdots,\tilde{\zeta}_{piK}\right)
\end{align}
where $\tilde{\zeta}_{pik} = ({\phi_{pk} \tilde{\theta}_{ki}})/{\sum_{k=1}^K \phi_{pk} \tilde{\theta}_{ki}}$.
  Therefore, S$\gamma\Gamma$-PFA has almost the same MCMC inference as the focused topic model (FTM) using the IBP compound Dirichlet priors \citep{FocusTopic}.
 Note that (\ref{eq:FTM1}) and (\ref{eq:FTM2}) are actually used in the FTM to infer $r_k$ and $z_{ki}$ \citep{FocusTopic} without giving explicit explanations under the multinomial-Dirichlet construction, however, we show that both equations naturally arise under S$\gamma\Gamma$-PFA under the constraint that $p_k=0.5$. In this sense, S$\gamma\Gamma$-PFA provides justifications for the inference in \citet{FocusTopic}.

\vspace{-0mm}
\section{Example Results and Discussions}\label{sec:experiments}
\vspace{-0mm}

We consider the JACM\footnote{http://www.cs.princeton.edu/$\sim$blei/downloads/} and PsyRev\footnote{\label{footnote}psiexp.ss.uci.edu/research/programs$\_$data/toolbox.htm} datasets, restricting the vocabulary to terms that occur in
five or more documents in each corpus. The JACM includes 536 abstracts of the Journal of the ACM from 1987 to 2004, with 1,539 unique terms and 68,055 total word counts; the PsyRev includes 1281 abstracts from
Psychological Review from 1967 to 2003, with 2,566 unique terms and 71,279 total word counts. As a comparison, the stopwords are kept in JACM and removed in PsyRev. We obtain similar results on other document corpora such as the NIPS corpus\footnote{http://cs.nyu.edu/$\sim$roweis/data.html}. We focus on these two datasets for detailed comparison.

For each corpus, we randomly select $80\%$ of the words from each document to form a training matrix $\Tmat$, holding out the remaining $20\%$ to form a testing matrix $\Ymat=\Xmat-\Tmat$.  We factorize $\Tmat$ as $\Tmat \sim \mbox{Pois}(\Phimat\Thetamat)$ and calculate the held-out per-word perplexity as
%\beq
\[
\exp\left( - \frac{1}{y_{\cdotv\cdotv}}\sum_{p=1}^{P}\sum_{i=1}^N y_{pi} \log \frac{\sum_{s=1}^S \sum_{k=1}^K \phi^s_{pk}\theta^s_{ki}}{\sum_{s=1}^S \sum_{p=1}^P\sum_{k=1}^K \phi^s_{pk}\theta^s_{ki}}\right)\nonumber
\]
%\eeq
where $S$ is the total number of collected samples, $y_{\cdotv\cdotv}=\sum_{p=1}^{P}\sum_{i=1}^N y_{pi}$ and $y_{pi}=\Ymat(p,i)$. The final results are based on the average of five random training/testing partitions. We consider 2500 MCMC iterations, with the first 1000 samples discarded and every sample per five iterations collected afterwards.
The performance measure is similar to those used in \cite{AsuWelSmy2009a,wallach09,VBHDP}\cite{}.

As discussed in Sec. \ref{Sec:RelatedModel} and shown in Table \ref{Tab:Relationships}, NMF \citep{NMF} is a special case of $\Gamma$-PFA; GaP \citep{CannyGaP} is a special case of $\beta\Gamma$-PFA; and in terms of inference, Dir-PFA is equivalent to LDA \citep{LDA,OnlineLDA}; and S$\gamma\Gamma$-PFA is closely related to FTM \citep{FocusTopic}.
Therefore, we are able to compare all these algorithms with $\beta\gamma\Gamma$-PFA under the same PFA framework, all with MCMC inference.

We set the priors of $\Gamma$-PFA as $a_\phi=a_\theta =1.01$, $b_\phi = 10^{-6}$, $g_k = 10^6$ and the prior of $\beta\Gamma$-PFA as $r_k =1.1$; these settings closely follow those that lead to the EM algorithms of NMF and GaP, respectively. We find that $\Gamma$- and $\beta\Gamma$-PFAs in these forms generally yield better prediction performance than their EM counterparts, thus we report the results of $\Gamma$- and $\beta\Gamma$-PFAs under these prior settings.
We set the prior of Dir-PFA as $a_\theta=50/K$, following the suggestion of the topic model toolbox$^2$ provided for \citet{FindSciTopic}. The parameters of $\beta\gamma\Gamma$-PFA are set as $c=c_0=r_0=\gamma=\alpha=1$. In this case, $\nu^+_{\epsilon\text{BP}}=\pi/\sin(\pi\epsilon)\approx \epsilon^{-1}$ for a small $\epsilon$, thus we preset a large upper-bound $K_{\max}$ and let $\epsilon=1/K_{\max}$.
The parameters of S$\gamma\Gamma$-PFA are set as $c=r_0=1$ and $\epsilon=1/K_{\max}$. The stepsize in (\ref{eq:randomMH}) is initialized as $\mu=0.01$ and is adaptively adjusted to maintain an acceptance rate between $25\%$ and $50\%$. For all the algorithms, $\Phimat$ and $\Thetamat$ are preset with random values. Under the above settings, it costs about 1.5 seconds per iteration for $\beta\gamma\Gamma$-PFA on the PsyRev corpus using a 2.67 GHz PC. %, with about $?\%$ of the computational time spent on sampling $x_{pik}$ with (\ref{eq:MultPois}).

\begin{figure}[!tb]
\begin{center}
\includegraphics[width=68mm]{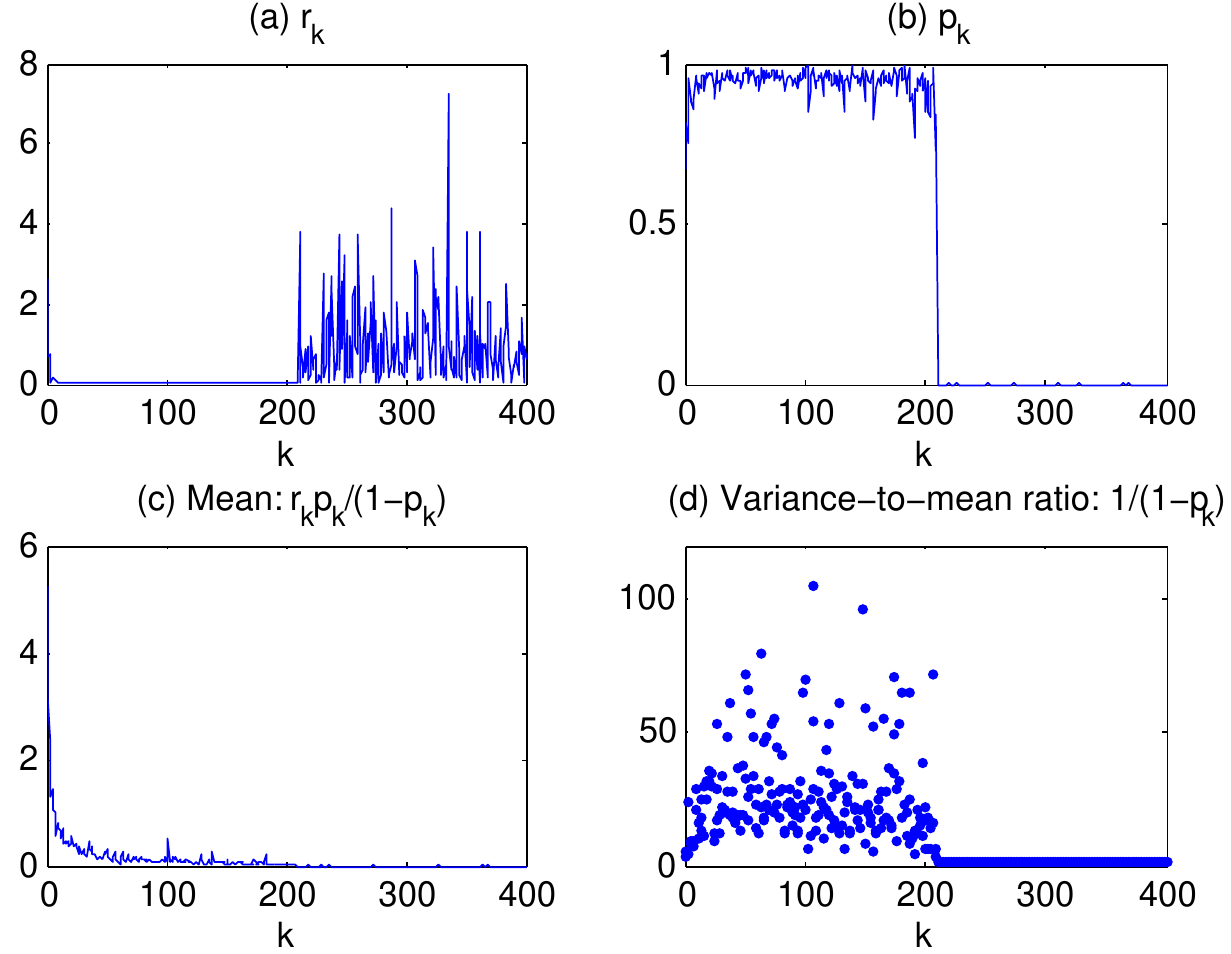}
\end{center}
\vspace{-3.8mm}
\caption{\small\label{fig:rkpk}
Inferred $r_k$, $p_k$, mean $r_kp_k/(1-p_k)$ and variance-to-mean ratio $1/(1-p_k)$ for each factor by $\beta\gamma\Gamma$-PFA with $a_\phi=0.05$. The factors are shown in decreasing order based on the total number of word counts assigned to them. Of the $K_{\max}=400$ possible factors, there are 209 active factors assigned nonzero counts. The results are obtained on the training count matrix of the PsyRev corpus based on the last MCMC iteration.\vspace{-3.8mm}
}
\end{figure}

Figure \ref{fig:rkpk} shows the inferred $r_k$ and $p_k$, and the inferred mean $r_kp_k/(1-p_k)$ and variance-to-mean ratio (VMR) $1/(1-p_k)$ for each latent factor using the $\beta\gamma\Gamma$-PFA algorithm on the PsyRev corpus. Of the $K=400$ possible factors, there are 209 active factors assigned nonzero counts. There is a sharp transition between the active and nonactive factors for the values of $r_k$ and $p_k$. The reason is that for nonactive factors, $p_k$ and $r_k$ are drawn from (\ref{eq:p_k_post}) and (\ref{eq:r_k0}), respectively, and thus $p_k$ with mean $c\epsilon/(c+Nr_k)$ is close to zero and $r_k$ is approximately drawn from its prior $\mbox{Gamma}(c_0r_0,1/c_0)$;  for active factors, the model adjusts the negative binomial distribution with both $p_k$ and $r_k$ to fit the data, and $r_k$ would be close to zero and $p_k$ would be close to one for an active factor with a small mean and a large VMR, as is often the case for both corpora considered.

We find that the first few dominant factors correspond to common topics popular both across and inside documents. For example, the first two most dominant topics are  characterized by ``proposed, evidence, data, discussed, experiments'' and ``model, models, effects, response, predictions''
in PysRev where the stop words are removed; the first two dominant topics in JACM, where the stop words are kept, are characterized by ``the, of, a, is, in'' and
``we, of, in, and, the''. These top factors generally have  large means and small VMRs.  The remaining topics have a diverse range of means and VMRs. A PsyRev factor with prominent words ``masking, visual, stimulus, metacontrast'' and a JACM factor with prominent words
``local, consistency, finite, constraint''
are example topics with large mean and large VMR. A PsyRev factor
``rivalry, binocular, monocular, existence''
 and a JACM factor
 ``search, binary, tree, nodes''
 are example topics with small mean and large VMR.
  Therefore, the $\beta\gamma\Gamma$-PFA captures topics with distinct characteristics by adjusting the negative binomial parameters $r_k$ and $p_k$, and the characteristics of these inferred parameters may assist in factor/topic interpretation.
Note that conventional topic models are susceptible to stop words, in that they may produce topics that are not readily interpretable if stop words are not removed \citep{nCRP}. Our results show that when stop words are present, $\beta\gamma\Gamma$-PFA usually absorbs them into a few dominant topics with large mean and small VMR and the remaining topics are  easily interpretable.

\begin{figure}[!tb]
\begin{center}
\includegraphics[width=38mm]{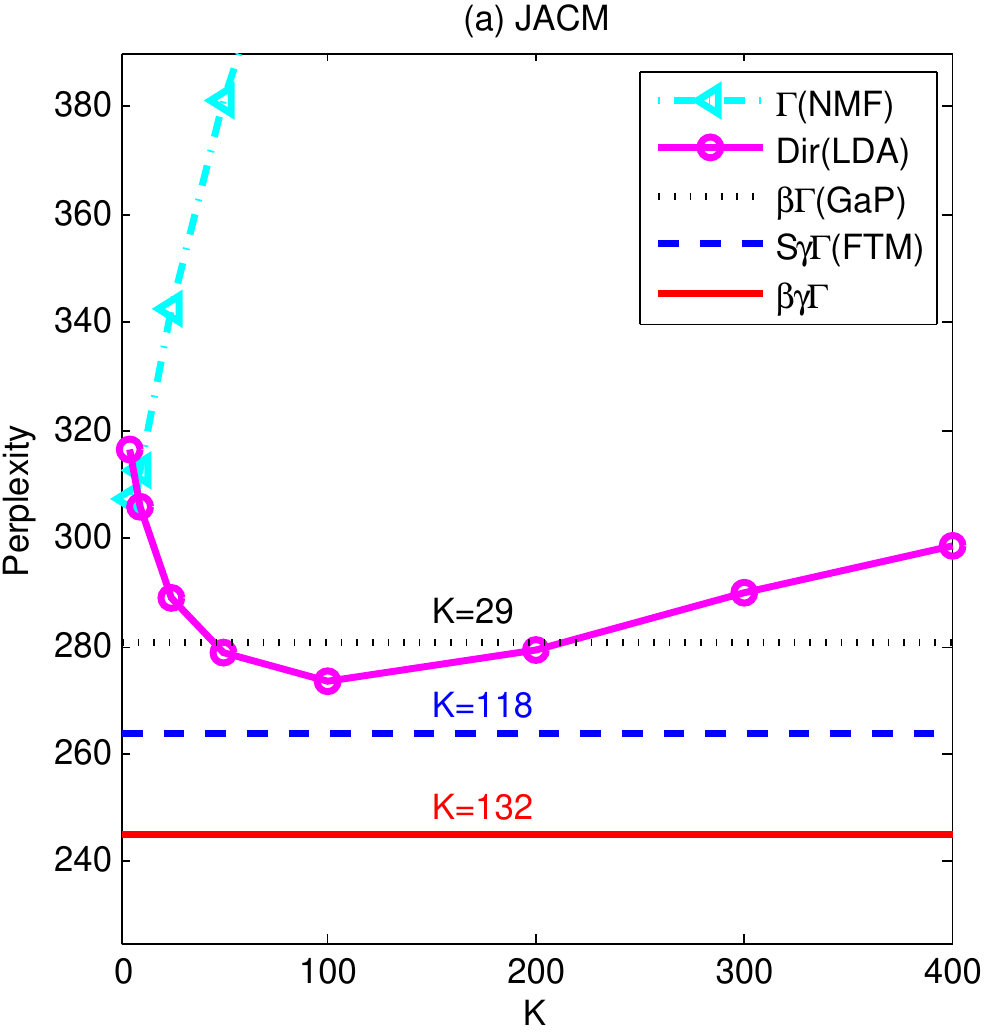}
\includegraphics[width=38mm]{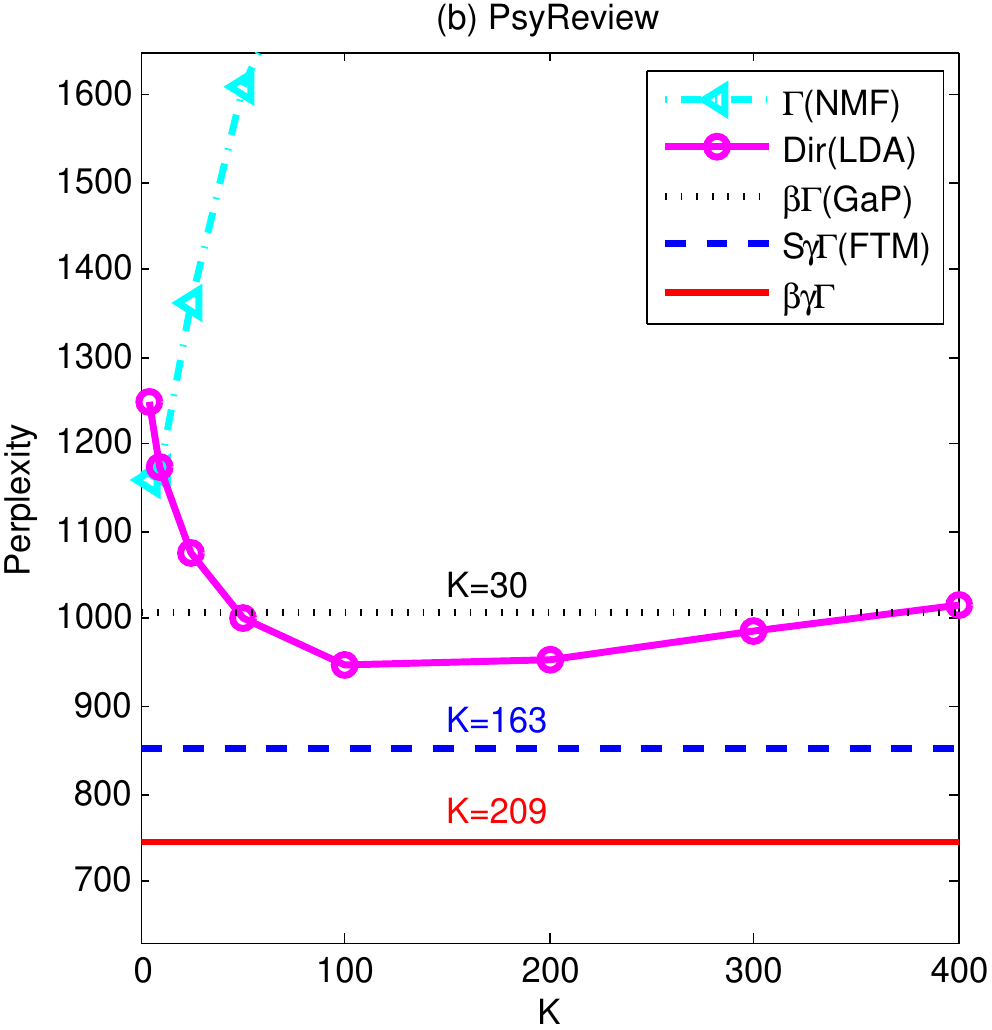}
\end{center}
\vspace{-3.9mm}
\caption{\small\label{fig:K}
Per-word perplexities on the test count matrix for (a) JACM and (b) PsyRev with $a_\phi = 0.05$. The results of $\Gamma$- and Dir-PFAs are a function of $K$. $\beta\Gamma\text{-}$, S$\gamma\Gamma$- and $\beta\gamma\Gamma$-PFAs all automatically infer the number of active factors, with the number at the the last MCMC iteration shown on top of the corresponding lines.
\vspace{-3.0mm}
}
\end{figure}

Figure \ref{fig:K} shows the performance comparison for both corpora with the factor loading (topic) Dirichlet prior  set as $a_\phi = 0.05$. In both $\Gamma$- and Dir-PFAs, the number of factors $K$ is a tuning parameter, and as $K$ increases, $\Gamma$-PFA quickly overfits the training data and Dir-PFA shows signs of ovefitting around $K=100$. In $\beta\Gamma\text{-}$, S$\gamma\Gamma$- and $\beta\gamma\Gamma$-PFAs, the number of factors is upper bounded by $K_{\max}=400$ and an appropriate $K$ is automatically inferred. As shown in Fig. \ref{fig:K}, $\beta\gamma\Gamma$-PFA produces the smallest held out perplexity, followed by S$\gamma\Gamma$- and $\beta\Gamma$-PFAs, and their inferred sizes of $K$ at the last MCMC iteration are 132, 118 and 29 for JACM and 209, 163 and 30 for PsyRev, respectively.

\begin{figure}[!tb]
\begin{center}
\includegraphics[width=38mm]{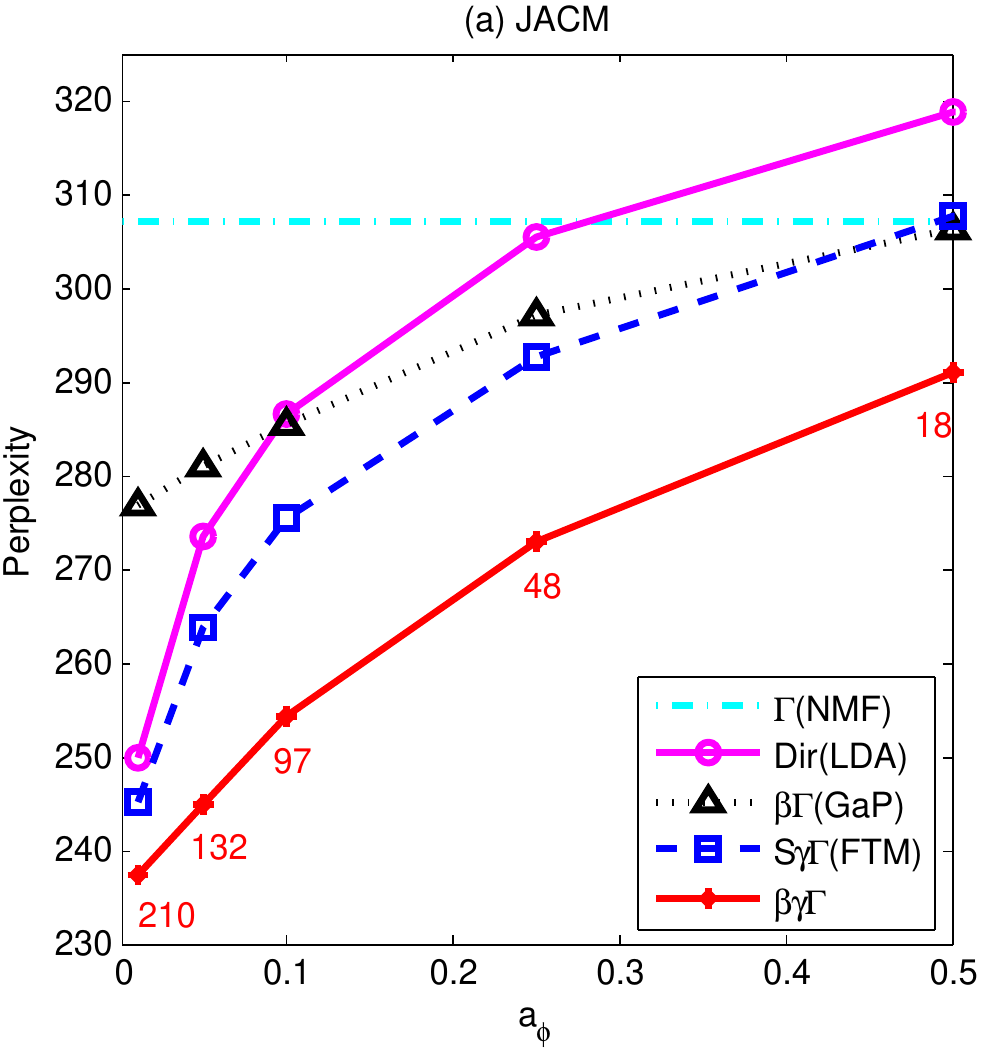}
\includegraphics[width=38mm]{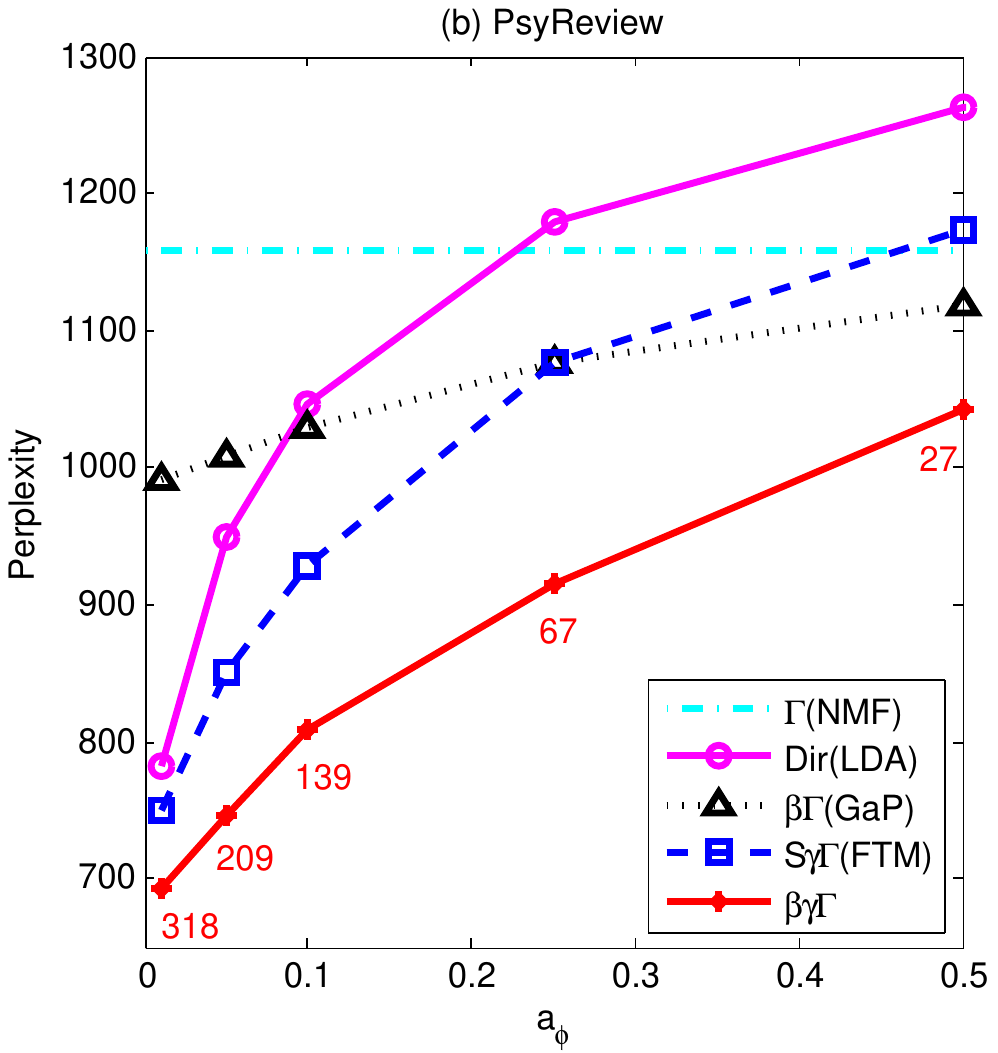}
\end{center}
\vspace{-3.9mm}
\caption{\small\label{fig:phi}
Per-word perplexities on the test count matrix of (a) JACM and (b) PsyRev as a function of the factor loading (topic) Dirichlet prior $a_\phi \in\{0.01, 0.05, 0.1,  0.25, 0.5\}$. The results of $\Gamma$- and Dir-PFAs are shown with the best settings of $K$ under each $a_\phi$. The number of active factors under each $a_\phi$ are all automatically inferred by $\beta\Gamma$-, S$\gamma\Gamma$- and $\beta\gamma\Gamma$-PFAs. The number of active factors inferred by $\beta\gamma\Gamma$-PFA at the last MCMC interaction are shown under the corresponding points.
\vspace{-3.0mm}
}
\end{figure}
Figure \ref{fig:phi} shows the performance of these algorithms as a function of $a_\phi$.  For $\Gamma$-PFA, $a_\phi$ is fixed. For Dir-, $\beta\Gamma$-, S$\gamma\Gamma$- and $\beta\gamma\Gamma$-PFAs, $a_\phi$ influences the inferred sizes of $K$ and the accuracies of held-out predictions. We find that a smaller $a_\phi$ generally supports a larger $K$, with better held-out prediction. However, if $a_\phi$ is too small it leads to overly specialized factor loadings (topics), that concentrate only on few terms. As shown in \ref{fig:phi}, $\beta\gamma\Gamma$-PFA yields the best results under each $a_\phi$ and it automatically infers the sizes of $K$ as a function of $a_\phi$.

\vspace{-1mm}
\section{Conclusions}\label{sec:conclusions}
\vspace{-1mm}
A beta-negative binomial (BNB) process, which leads to a beta-gamma-Poisson process, is proposed for modeling multivariate count data. The BNB process is augmented into a beta-gamma-gamma-Poisson hierarchical structure and applied as a nonparametric Bayesian prior for Poisson factor analysis (PFA), an infinite  discrete latent variable model. A finite approximation to the beta process L\'{e}vy random measure is proposed for convenient implementation. Efficient MCMC inference is performed by exploiting the relationships between the beta, gamma, Poisson, negative binomial, multinomial and Dirichlet distributions. Connections to previous models are revealed with detailed analysis. Model properties are discussed, and example results are presented on document count matrix factorization. Results demonstrate that by modeling latent factors with negative binomial distributions whose mean and variance are both learned, the proposed $\beta\gamma\Gamma$-PFA is well suited for topic modeling, defined quantitatively via perplexity calculations and more subjectively by capturing both common and specific aspects  of a document corpus.

\newpage
\subsubsection*{Acknowledgements}
The research reported here was supported by AFOSR, ARO, DARPA, DOE, NGA, and ONR.

\bibliographystyle{plainnat}
\bibliography{PGFA,HBP_AISTATS2011}

\end{document}